\documentclass[conference]{IEEEtran}
\IEEEoverridecommandlockouts
\usepackage{cite}
\usepackage{subcaption}
\usepackage{float}
\usepackage{amsmath,amssymb,amsfonts}
\usepackage{algorithmic}
\usepackage{graphicx}
\usepackage{textcomp}
\usepackage{xcolor}
\usepackage{orcidlink}
\usepackage{hyperref}
\def\BibTeX{{\rm B\kern-.05em{\sc i\kern-.025em b}\kern-.08em
    T\kern-.1667em\lower.7ex\hbox{E}\kern-.125emX}}

\title{TGO-IV: Developmental Topology Observatory}
\author{
\IEEEauthorblockN{Kaustubh Kapil\orcidlink{0009-0000-4918-8452} and Kishor P. Upla\orcidlink{0000-0001-6306-0682}}
\IEEEauthorblockA{
\textbf{\textit{Neuromorphic Intelligence Research Collective}}\\
Department of Electronics Engineering\\
Sardar Vallabhai National Institute of Technology (SVNIT), Surat, India\\
kaustubhkapil2006@gmail, kishorupla@gmail.com
}
}
\begin{document}

\maketitle

\begin{abstract}
Transformers have had a profound impact on the world of language processing and computer vision. As efforts to answer the million-dollar question of ``How does a Transformer learn?" have been increasing, existing interpretability studies primarily analyze representations at isolated layers or the network as a whole, while the developmental evolution of individual representations and its manifolds across transformer layers remains underexplored. With this work, we aim at providing a comprehensive analysis of the evolution of representations as the representation point cloud transforms across the layers; thereby attempting to isolate layers or establish a trend which comes closer to justifying how and when raw input representations evolve into task-relevant feature representations. Thus, Transformer Geometry Observatory-TGO-IV introduces a topological framework for analysing the evolution of Transformer representations through the lens of Persistent Homology. Rather than studying local geometric properties alone, TGO-IV constructs Vietoris--Rips simplicial complexes from token-level representation point clouds and investigates the evolution of their persistent topological signatures across Transformer layers. The proposed framework comprises complementary topological observatories including Persistence Diagrams, Barcode Diagrams, Betti Curves, Persistence Landscapes, Bottleneck Distance, and Wasserstein Distance, enabling a comprehensive analysis of how the global topology of representation point clouds develops throughout the forward pass.
\end{abstract}

\section{Introduction}
Perhaps, one of the most effective methods to establish true feature correlation and retention of global context was demonstrated by \emph{Transformers}~\cite{attention,vit}. With the empirical success however, came the question of how transformers learn this well. The first key instincts was to try and understand the transformer's working dynamics in two prongs, representational dynamics along the forward path and loss landscape while back propagation. Most of the studies involving transformers and their applications rarely treat it like a black-box; they instead try to isolate the layers or individual computational blocks and perform spectral analyses on the representations obtained from each layers, common examples include studies like ~\cite{svcca},~\cite{cka},~\cite{geometricrepr}. While these studies do establish a solid background as to how to develop effective tests and understand how spectrum of various representations are linked, they still fail to capture continuous evolution of a representation throughout the entire black-box. 

\textbf{Transformer Geometry Observatory} is a multi-part framework which investigates the evolution of spectral dynamics by consolidating the multiple preexisting methods involving spectral, geometric, and topological methods. \textbf{TGO-I: Spectral Geometry Observatory}~\cite{tgoi} showed that during training, the covariance spectrum gradually grows along with rising Effective Rank, Stable Rank, and Spectral Entropy, indicating that the network is constantly exploring new representational directions. Later, \textbf{TGO-II: Representation Geometry Observatory}~\cite{tgoii} showed that while representational similarity between layers gradually declines, the inherent dimensionality of Transformer representations rises throughout optimisation. The Manifold growth Hypothesis, which postulates that the observed covariance growth results from the ongoing expansion of the underlying representation manifold rather than token diversification alone, was inspired by these data. In order to solidify these claims the \textbf{TGO-III: Semantic Geometry Observatory}~\cite{tgoiii} was introduced which further strengthened the semantic expansion and transition zone hypotheses, by demonstrating how this expanding representation manifold is gradually arranged into semantic structures that are more discriminative. By linking covariance evolution, manifold expansion, and semantic organization under a single geometric framework, this approach creates a cohesive picture of Transformer representation learning.

However, despite these observations, one important question remains unanswered: does the representation manifold itself preserve its global topological structure throughout the forward pass, or does it undergo fundamental topological transitions? While covariance spectra, intrinsic dimensionality, Jacobian fields, and semantic observatories characterize the local geometry of representations, they are inherently insensitive to global topological invariants. Two manifolds may possess vastly different geometric properties while remaining topologically equivalent, or conversely exhibit identical local geometry despite undergoing substantial changes in global connectivity. Consequently, geometric analyses alone cannot determine whether Transformer layers preserve, create, merge, or eliminate fundamental topological structures during representation development. To address this limitation, we introduce TGO-IV: Developmental Topology Observatory, which extends the Transformer Geometry Observatory framework into Topological Data Analysis (TDA). Rather than analyzing the unknown representation manifold directly, TGO-IV constructs Vietoris–Rips simplicial complexes from token-level representation point clouds and studies their persistent homology across Transformer layers. Through Persistence Diagrams, Barcode Diagrams, Betti Curves, Bottleneck Distances, and Wasserstein Distances, we quantify how the topology of these simplicial-complex approximations evolves throughout the network, providing the first topological observatory of representation development in Vision Transformers. With that in mind, the key contributions of TGO-IV can be summarized as:
\begin{itemize}
    \item To introduce the first topological observatory for Transformer representations, extending the Transformer Geometry Observatory framework from spectral and geometric analysis to Topological Data Analysis (TDA) by studying the evolution of persistent topological structures throughout the forward pass.
\item To formulate a topology-centric analysis framework that constructs Vietoris--Rips simplicial complexes from token-level representation point clouds and systematically characterizes their evolution using persistent homology, enabling layer-wise investigation of global representation topology.
\item To propose a suite of complementary topological observatories consisting of Persistence Diagrams, Barcode Diagrams, Betti Curves, Persistence Landscapes, Bottleneck Distance, and Wasserstein Distance to quantify the birth, persistence, and disappearance of topological features across Transformer layers.
\item Finally, to investigate the developmental evolution of representation topology by analyzing how persistent topological signatures emerge, transform, and stabilize throughout the network, providing new insights beyond conventional geometric and spectral analyses.

\end{itemize}
\subsection{Research Questions}

With this study, we aim to answer the following questions:
\begin{itemize}
    \item \textbf{RQ1: Does representation topology evolve systematically across Transformer depth?} Present studies limit analyses to CKA~\cite{cka}, SVCCA~\cite{svcca}, and Intrinsic Dimensions~\cite{twonn}. It is scarcely explored on how the representation manifold may evolve topologically as layers progress and training matures.
    \item \textbf{RQ2: Are topological transitions concentrated in specific Transformer layers?} One of the fundamental questions that has stuck with the community is whether evolution of representations, data manifolds, or semantic meaning takes place discretely in unique layers or is it a continuous process. Same can be said about topological features - are they discrete stepwise developments or continuous feature emergence?
    \item \textbf{RQ3: Can topological analyses be used to understand how a transformer works, how its representational transformations (dot products) tend to work?} This has been the most important question that TGO aims to answer - how does the transformer do what it does. TGO-IV does not clearly answer that but does aim to bring us closer to understanding the representational evolution taking place within the transformer. Topological analyses give us an insight beyond geometric arrangements, helping in analyzing feature emergence and persistence.
\end{itemize}
The complete implementation of TGO-IV, observatory pipelines,
and generated artifacts are publicly available at:
\href{https://github.com/KaustubhKapil/Transformer_Geometry_Observatory_Part-4}{GitHub Repository}.

\begin{figure*}[h]
    \centering
    \includegraphics[width=0.9\linewidth]{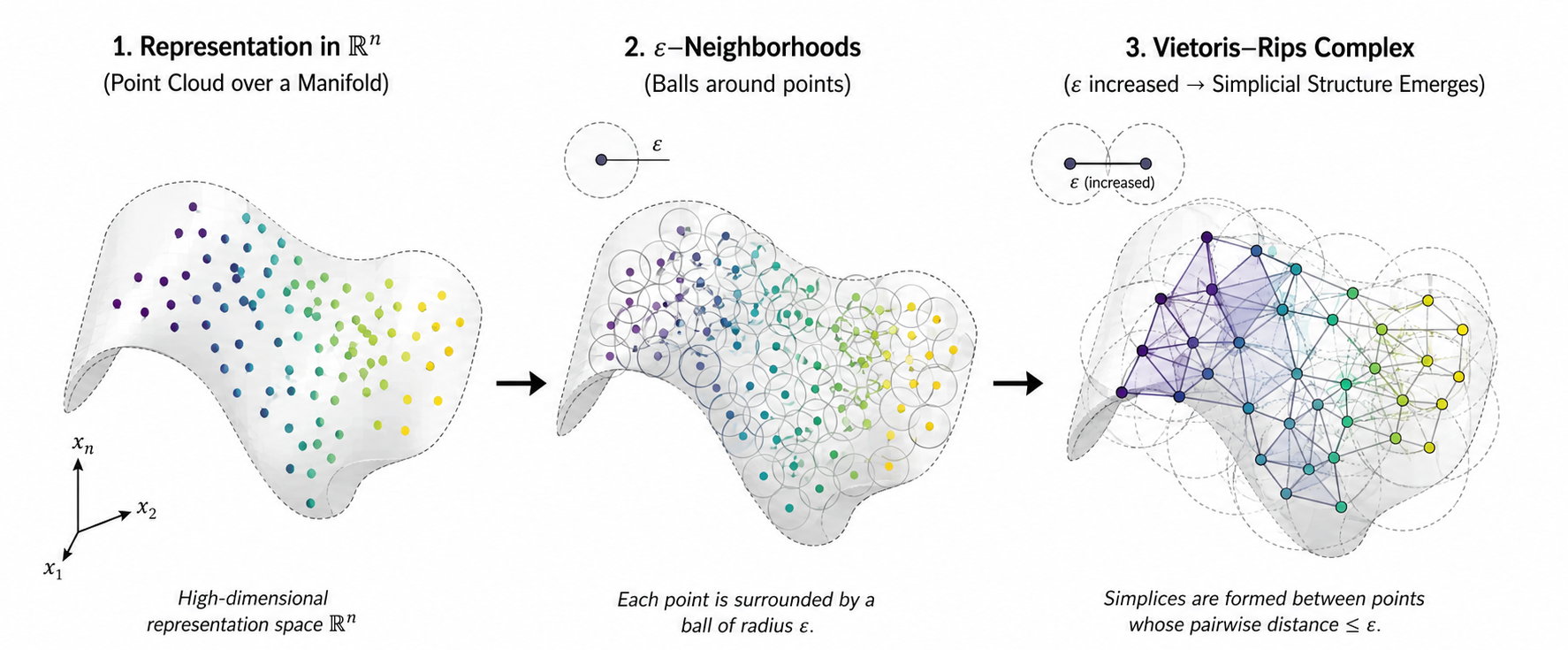}
    \caption{Methodology used to construct simplecial complexes using the representations as points in a multidimensional space.}
    \label{fig:rips}
\end{figure*}

\section{Methodology}
\

The objective of TGO-IV is to analyze the developmental evolution of Transformer representations from a topological perspective. Unlike conventional visualization approaches that project high-dimensional representations into two or three dimensions using techniques such as PCA, t-SNE, or UMAP, TGO-IV performs all analyses directly within the original representation space. Although low-dimensional projections provide intuitive visualizations, they inevitably distort pairwise distances and neighbourhood relationships, which form the basis of Topological Data Analysis (TDA). Since persistent topology is fundamentally determined by the metric relationships between points, all analyses are performed in the original embedding space $\mathbb{R}^{d}$.

For a Vision Transformer, every encoder block produces a collection of token embeddings

\begin{equation}
X^{(l)}
=
\{x_1^{(l)},x_2^{(l)},\ldots,x_N^{(l)}\},
\qquad
x_i^{(l)}\in\mathbb{R}^{d},
\end{equation}
where $l$ denotes the Transformer layer, $N$ is the number of tokens, and $d$ is the embedding dimension. Rather than analysing each embedding independently, TGO-IV interprets the complete collection of token embeddings at every layer as a representation point cloud. As representations propagate through successive Transformer layers, this point cloud continuously deforms, expands, contracts, and reorganizes within the embedding space.

Although the representation point cloud captures the geometric arrangement of token embeddings, topology cannot be extracted directly from a finite collection of points. Instead, it is necessary to construct a topological object whose connectivity approximates the underlying structure of the point cloud. To achieve this, TGO-IV constructs a Vietoris--Rips simplicial complex from every layer-wise representation point cloud. Given a filtration radius $\epsilon$, two points are connected whenever
\begin{equation}
\|x_i-x_j\|\leq\epsilon.
\end{equation}

Higher-dimensional simplices are subsequently formed whenever all lower-dimensional faces are present. Increasing the filtration radius progressively produces a nested family of simplicial complexes,
\begin{equation}
VR(\epsilon_1)
\subseteq
VR(\epsilon_2)
\subseteq
\cdots
\subseteq
VR(\epsilon_m),
\qquad
\epsilon_1<\epsilon_2<\cdots<\epsilon_m,
\end{equation}
which approximates the topology of the representation point cloud across multiple spatial scales.

Instead of analyzing a single simplicial complex, Persistent Homology tracks how topological structures evolve throughout the entire filtration. As $\epsilon$ increases, connected components merge, loops emerge and disappear, and higher-dimensional cavities are created and filled. Every topological feature is therefore associated with a birth scale and a death scale, providing a multiscale description of the representation topology rather than a topology at a single resolution.

Since persistent topology cannot be interpreted directly from the filtration itself, TGO-IV characterizes it using a collection of complementary topological observables. Persistence Diagrams and Barcode Diagrams summarize the lifetime of every topological feature, Betti Curves quantify the evolution of connected components and higher-dimensional structures throughout the filtration, Persistence Landscapes provide functional representations suitable for statistical comparison, while Bottleneck and Wasserstein distances quantify the similarity between persistent topological signatures obtained from different Transformer layers or different stages of training. Together, these observatories provide a comprehensive description of how the topology of Transformer representation point clouds develops throughout the forward pass. Fig~\ref{fig:rips} gives a visual cue to how the simplicial complex was created.

\emph{\textbf{METHODOLOGICAL CONSIDERATIONS:} While previous TGO experiments are conducted along the conventional guidelines without any radical assumptions, TGO-IV works on an underlying assumption. Simplicial complexes can be used to approximate the shape of any manifold given that the manifold is sufficiently smooth, there are sufficient number of samples and that the sample neighbourhood is relatively populous (HIGH DENSITY). In contrast, each layer of a ViT-Small/16 provides only 197 token representations in a 384-dimensional embedding space, resulting in an extremely sparse sampling of the underlying representation geometry. Consequently, the Vietoris--Rips complexes constructed in this work should not be interpreted as faithful reconstructions of the true representation manifold. Instead, they are treated as topological approximations whose persistent signatures may nevertheless provide useful insight into the developmental evolution of Transformer representations. Therefore, TGO-IV does not claim that the observed topological structures correspond directly to the intrinsic topology of the underlying representation manifold. Rather, the observatory investigates whether persistent topological descriptors exhibit consistent and meaningful developmental trends across Transformer layers and throughout training.}

\section{Observables}

\subsubsection{Persistence Diagram}

Persistence Diagrams provide a compact summary of the persistent homology of a filtration. Every topological feature is represented as a point whose coordinates correspond to its birth and death filtration values,
\begin{equation}
(b_i,d_i),
\end{equation}
where $b_i$ denotes the filtration value at which the feature first appears and $d_i$ denotes the value at which it disappears. Features lying farther from the diagonal possess longer persistence and are generally regarded as more significant, whereas points close to the diagonal typically correspond to topological noise. Persistence diagrams therefore provide a concise representation of the lifetime of every topological feature throughout the filtration.

\subsubsection{Barcode Diagram}

Barcode Diagrams visualize persistent homology by representing every topological feature as a horizontal interval extending from its birth to its death filtration value. Longer bars correspond to persistent topological structures, whereas shorter bars generally indicate transient features. Barcode diagrams provide an intuitive visualization of how connected components, loops, and higher-dimensional cavities evolve throughout the filtration.

\subsubsection{Betti Curves}

While Persistence Diagrams summarize individual topological features, Betti Curves quantify the number of topological features present at every filtration scale. For homology dimension $k$, the Betti number, $\beta_k(\epsilon)$ denotes the number of $k$-dimensional topological features existing at filtration value $\epsilon$. In particular,

\begin{itemize}
\item $\beta_0$ measures connected components,
\item $\beta_1$ measures one-dimensional loops,
\item $\beta_2$ measures enclosed voids.
\end{itemize}
Tracking Betti Curves across Transformer layers provides insight into how global connectivity and higher-order structures evolve during representation learning.

\subsubsection{Persistence Landscape}

Persistence Landscapes transform Persistence Diagrams into functional representations that facilitate statistical comparison between different layers and training epochs. Instead of representing topology as discrete points, persistence landscapes encode persistence information as a sequence of piecewise-linear functions, enabling averaging, hypothesis testing, and quantitative comparison of persistent topological structures.

\subsubsection{Bottleneck Distance}

To quantify changes in topology between Transformer layers, TGO-IV computes the Bottleneck Distance between Persistence Diagrams. Given two persistence diagrams $D_1$ and $D_2$, the Bottleneck Distance is defined as
\begin{equation}
d_B(D_1,D_2)
=
\inf_{\gamma}
\sup_{x\in D_1}
\|x-\gamma(x)\|_{\infty},
\end{equation}
where $\gamma$ denotes a bijection between diagram points. The Bottleneck Distance measures the largest topological discrepancy between two persistence diagrams and therefore captures the most significant structural change occurring between successive Transformer layers.

\subsubsection{Wasserstein Distance}

Whereas the Bottleneck Distance is governed by the single largest topological difference, the Wasserstein Distance measures the cumulative discrepancy between two persistence diagrams. It is defined as
\begin{equation}
W_p(D_1,D_2)
=
\left(
\inf_{\gamma}
\sum_{x\in D_1}
\|x-\gamma(x)\|^p_{\infty}
\right)^{1/p}.
\end{equation}
Consequently, Wasserstein Distance is sensitive to the overall distribution of topological features rather than only the largest change, providing a complementary measure of topological evolution across Transformer layers.

\section{Experiments}

This section describes the experimental configuration used to investigate the topological evolution of Vision Transformer representations throughout training. A ViT-Small/16 model was trained on the ImageNet-100 dataset for 100 epochs using an NVIDIA Quadro RTX 6000 GPU. To ensure consistent observability, all layer-wise measurements were performed using a fixed validation analysis subset of 1000 images throughout training. In addition, a single designated validation image was tracked across layers to construct the persistent topological observables reported in TGO-IV. The following subsections describe the model configuration, analysis protocol, topological extraction pipeline, and training parameters used throughout the study.

\begin{figure*}[!t]
    \centering
    \includegraphics[width=0.8\linewidth]{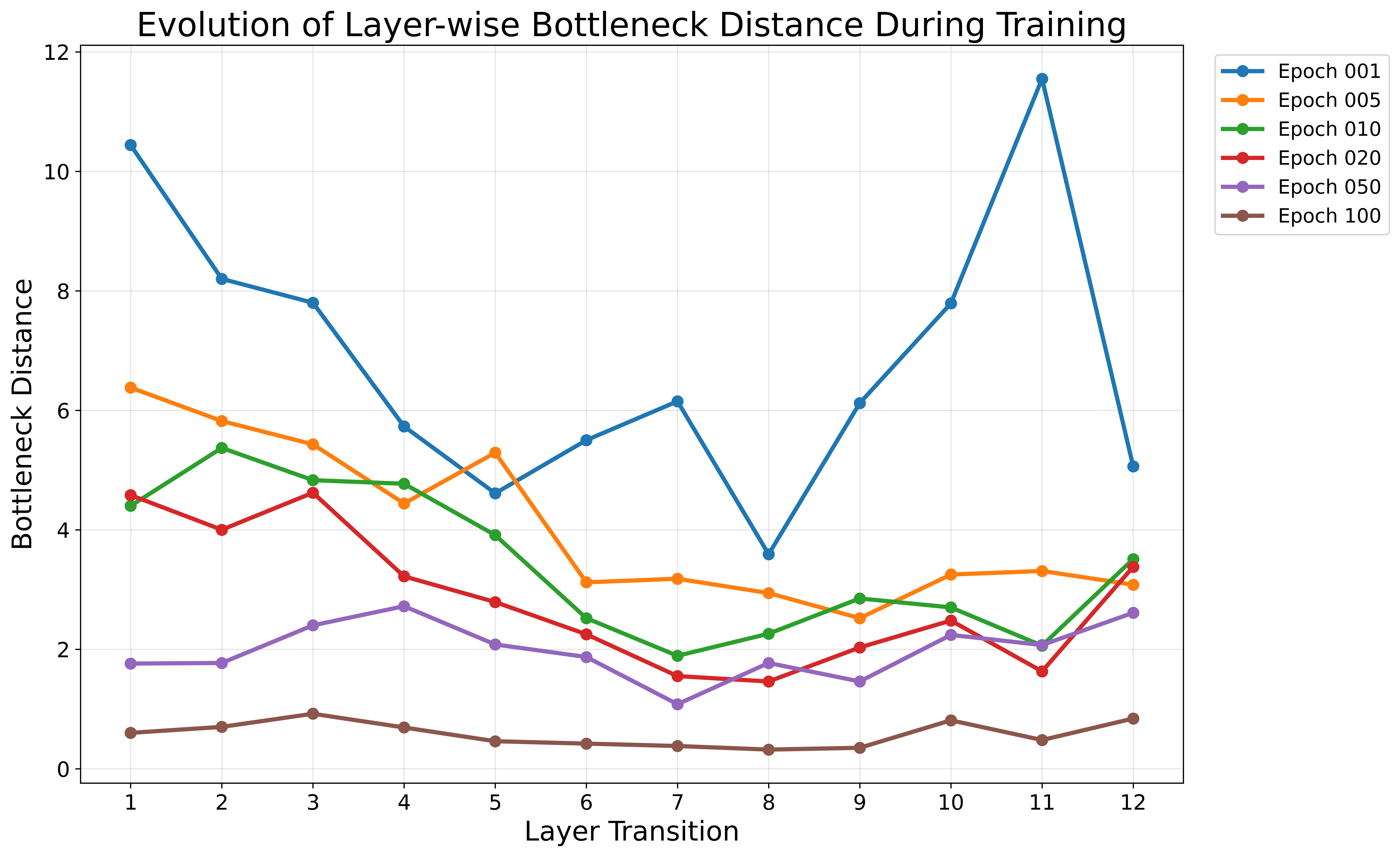}
    \caption{Evolution of the Bottleneck Distance between persistence diagrams of consecutive Transformer layers at representative training epochs. Larger values indicate greater topological dissimilarity between adjacent layer representations, while the overall decrease throughout training suggests progressively more similar persistent topological signatures across successive layers.}
    \label{fig:bottleneck_evolution}
\end{figure*}

\subsection{Model Configuration}

The model used throughout TGO-IV was ViT-Small/16. The architecture consists of a patch embedding layer followed by twelve Transformer encoder blocks and a final classification head operating on the CLS token representation.
Input images were partitioned into non-overlapping $16 \times 16$ patches and projected into a 384-dimensional embedding space. A learnable CLS token and positional embeddings were added prior to Transformer processing. The resulting token sequence was propagated through twelve encoder blocks composed of Multi-Head Self-Attention and Feed-Forward Network modules.
For an input resolution of $224 \times 224$, the model produces a sequence of 197 tokens, comprising 196 patch tokens and one CLS token.

\subsection{Analysis Protocol}

To preserve longitudinal consistency, all observables were computed using fixed evaluation subsets rather than training mini-batches. A validation subset of 1000 images was used for layer-wise population-level analysis, while a designated single validation image was used for detailed representation-topology tracking across depth.

For every monitored layer, the token embeddings were treated as a point cloud in $\mathbb{R}^{d}$. For a single image, this yields a layer-wise representation matrix
\begin{equation}
\mathbf{X}^{(l)} \in \mathbb{R}^{T \times D},
\end{equation}
where $T$ denotes the number of tokens and $D$ denotes the embedding dimension. In the ViT-Small/16 setting, $T = 197$ and $D = 384$. This construction provides the input point cloud on which all topological analysis is performed.

\subsection{Topological Construction}

Topology is not measured directly from the raw embeddings. Instead, each layer-wise point cloud is converted into a Vietoris--Rips simplicial complex by introducing a filtration radius $\epsilon$ and connecting points whose pairwise distance satisfies
\begin{equation}
\|x_i - x_j\| \leq \epsilon.
\end{equation}
As $\epsilon$ increases, higher-dimensional simplices are formed whenever all lower-dimensional faces are present, producing a nested filtration
\begin{equation}
VR(\epsilon_1) \subseteq VR(\epsilon_2) \subseteq \cdots \subseteq VR(\epsilon_m),
\qquad
\epsilon_1 < \epsilon_2 < \cdots < \epsilon_m.
\end{equation}
Persistent Homology is then applied to this filtration to track the birth, persistence, and disappearance of connected components, loops, and higher-dimensional cavities. Since topology itself cannot be directly visualized from the raw point cloud, TGO-IV summarizes the evolving persistent structure using a set of complementary observables, namely Persistence Diagrams, Barcode Diagrams, Betti Curves, Persistence Landscapes, Bottleneck Distance, and Wasserstein Distance.

\subsection{Training Configuration}

The ViT-Small/16 model was trained using PyTorch with Automatic Mixed Precision (AMP) enabled. Optimization was performed with AdamW using a learning rate of $10^{-3}$, weight decay of $0.05$, label smoothing of $0.1$, and gradient clipping at $1.0$. A cosine annealing scheduler was used with a minimum learning rate of $10^{-6}$. The random seed was fixed to ensure reproducibility.

Model checkpoints were saved throughout training as both the best-performing checkpoint and the most recent checkpoint. At the end of each epoch, the fixed analysis subsets were processed through the network and the corresponding topological observables were computed. For selected epochs, namely 1, 5, 10, 20, 50, and 100, the observables were stored for longitudinal comparison, yielding a complete temporal record of the evolution of representation topology across both network depth and training time.

\section{Findings}

In this section, we shall explore, in detail, the observations of the TGO-IV Framework. As a reiertation to a previous consideration, 197 points are not sufficient to extract faithful manifold approximations of the representational landscape and hence we treat the simplicial complex and its topology as a separate measurement paradigm instead of working with manifold topology. The key observations of TGO-IV have been listed below.

\subsection{Bottleneck Distance Evolution}

The Bottleneck Distance quantifies the largest topological discrepancy between the persistence diagrams of consecutive Transformer layers. Larger values indicate substantial changes in the persistent topological signature between adjacent layers, whereas smaller values suggest increasingly similar topological structures. Figure~\ref{fig:bottleneck_evolution} illustrates the evolution of the Bottleneck Distance throughout training. At Epoch 1, large bottleneck distances are observed across most layer transitions, indicating considerable topological variation between successive representations. As training progresses, the overall magnitude decreases consistently, with Epoch 100 exhibiting substantially smaller distances throughout the network. Despite this global reduction, several localized peaks persist across training, particularly around transitions near Layers 3, 10, and 12. These transitions consistently exhibit comparatively larger topological changes than neighbouring layers, suggesting that specific stages of the Transformer continue to perform more pronounced topological reorganizations even after convergence. Overall, the decreasing Bottleneck Distance suggests that the persistent topological signatures of neighbouring layers become progressively more similar as optimization proceeds. Consistent with the methodological considerations discussed in Section~III, these observations should be interpreted as comparative changes in persistent topological signatures rather than exact measurements of the intrinsic topology of the underlying representation manifold.

\subsection{Betti Curve Evolution and Barcode evolution}

\begin{figure*}[!t]
    \centering
    \includegraphics[width=1.0\linewidth]{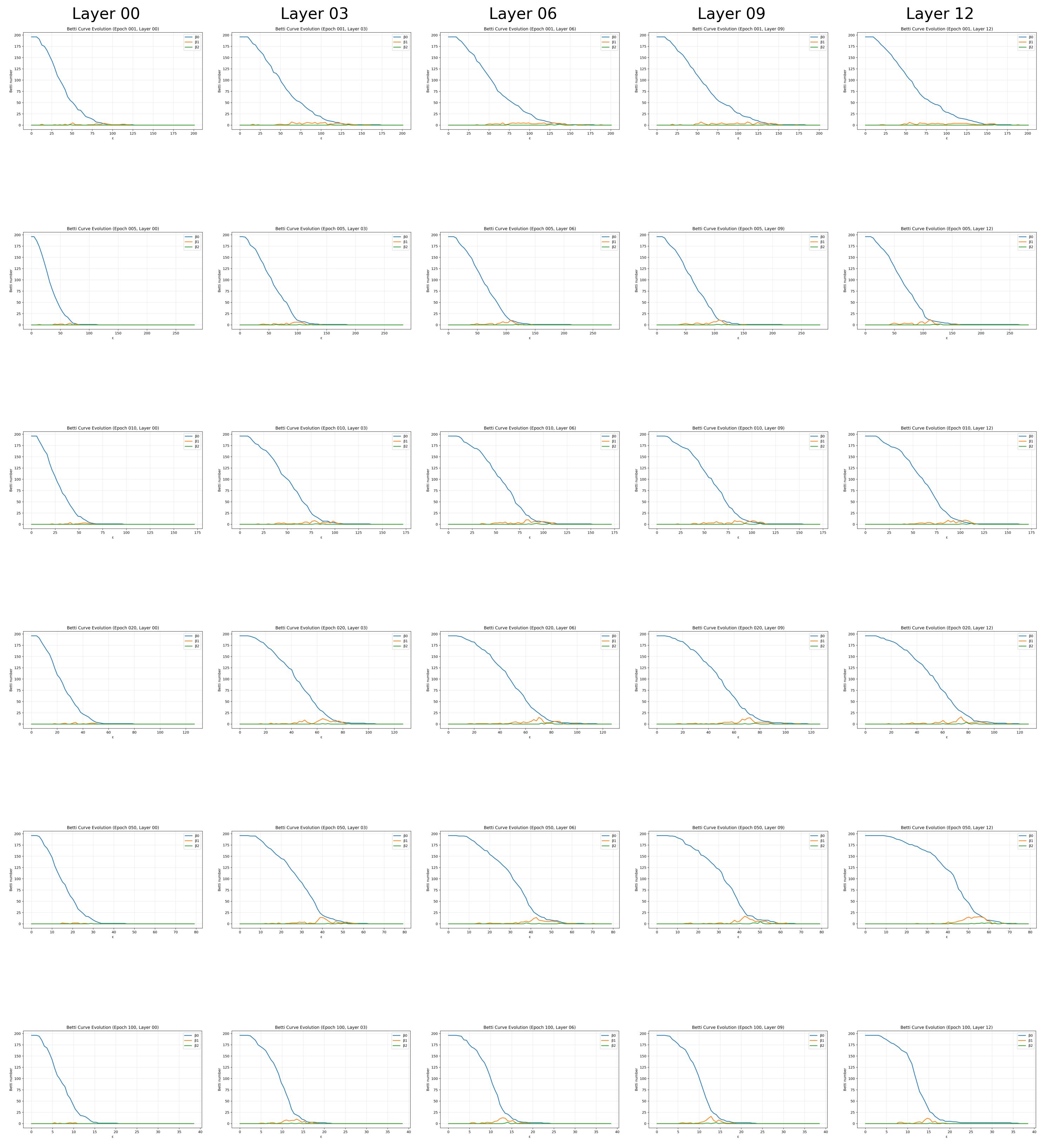}
    \caption{Evolution of the Betti curves for representative Transformer layers throughout training. The plots show the zeroth ($\beta_0$), first ($\beta_1$), and second ($\beta_2$) Betti numbers as functions of the filtration radius. Across all epochs, $\beta_0$ dominates the topology, while $\beta_1$ remains limited and $\beta_2$ is largely absent, indicating that the persistent topology is primarily characterized by the connectivity of the representation point cloud.}
    \label{fig:betti_curves}
\end{figure*}

Betti Curves provide a multiscale summary of the persistent topological features present throughout the filtration process. Figure~\ref{fig:betti_curves} illustrates the evolution of the Betti curves for representative Transformer layers at different stages of training. Across all layers and training epochs, the topology is dominated by the zeroth Betti number ($\beta_0$), corresponding to the number of connected components. At small filtration radii, each token initially forms an isolated component, resulting in a large $\beta_0$, which decreases monotonically as neighbouring points merge during the filtration. As training progresses, the $\beta_0$ curves consistently shift towards smaller filtration radii. This indicates that connected components merge earlier during the filtration, suggesting that the representation point cloud becomes progressively more cohesive as learning proceeds. This behaviour is observed consistently across all monitored Transformer layers. In contrast, the first Betti number ($\beta_1$), representing one-dimensional loops, remains comparatively small throughout training, exhibiting only localized peaks over a narrow range of filtration values. Furthermore, the second Betti number ($\beta_2$) remains nearly zero across all experiments, indicating the absence of persistent higher-dimensional cavities within the constructed simplicial complexes. Overall, the Betti Curves suggest that the dominant topological evolution throughout training is characterized by increasing connectivity of the representation point cloud, while higher-order topological structures remain relatively limited. Consistent with the methodological considerations discussed in Section~III, these observations should be interpreted as comparative changes in the persistent topology of the constructed simplicial complexes rather than exact measurements of the underlying representation manifold.

\begin{figure*}[!t]
    \centering
    \includegraphics[width=1.0\linewidth]{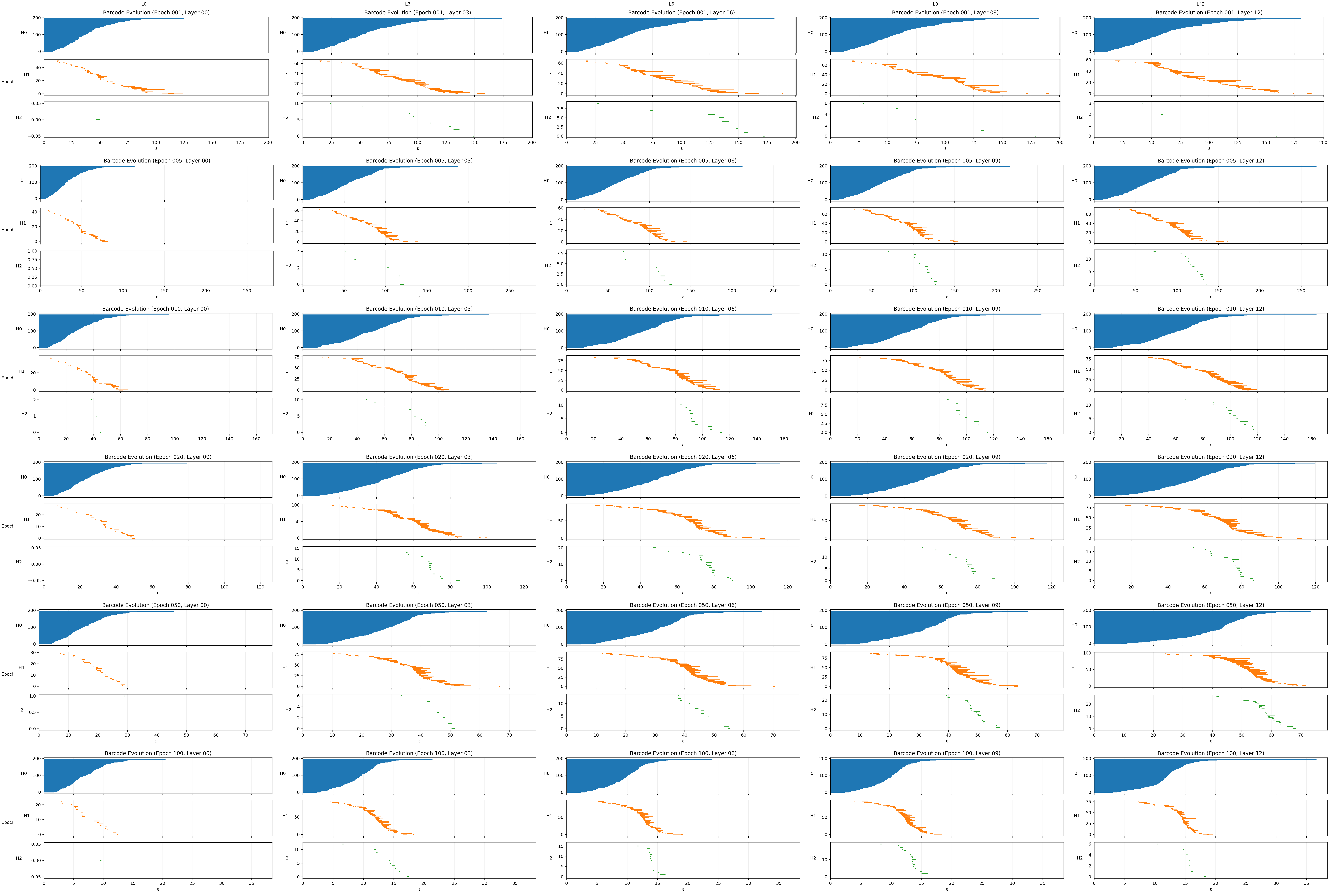}
\caption{Evolution of persistence barcode diagrams for representative Transformer layers throughout training. Each row corresponds to a training epoch, while each column represents a selected Transformer layer. Blue, orange, and green bars denote the persistent intervals of the zeroth ($H_0$), first ($H_1$), and second ($H_2$) homology groups, respectively. Across training, the barcodes indicate progressively shorter filtration ranges and increasingly stable persistent topological signatures, while higher-dimensional topological features remain comparatively sparse.}
    \label{fig:barcode}
\end{figure*}

Looking at the Barcode evolution as shown in Fig~\ref{fig:barcode} we can establish the coherence with the betti plots. We can summarize the findings of these plots as follows:
\begin{itemize}
    \item Across all layers and training epochs, the topology is overwhelmingly dominated by the zeroth homology group ($H_0$), indicating that the persistent topology is primarily governed by the connectivity of the representation point cloud.

    \item The $\beta_0$ curves and corresponding $H_0$ barcode intervals progressively contract towards smaller filtration radii throughout training, suggesting that connected components merge earlier as the learned representations become increasingly cohesive.

    \item The first homology group ($H_1$) remains comparatively sparse, with only a small number of short-lived persistent loops observed across a limited filtration range. No substantial increase in persistent loop structures is observed during training.

    \item The second homology group ($H_2$) is almost entirely absent throughout all layers and epochs, indicating that higher-dimensional cavities do not constitute a significant component of the persistent topology obtained from the representation point clouds.

    \item Although minor layer-wise variations are present, the overall shape of the Betti curves and barcode diagrams remains remarkably consistent across Transformer depth. The dominant developmental trend is instead observed across training epochs, where the filtration required to establish global connectivity decreases progressively.

    \item Collectively, the Betti curves and barcode diagrams suggest that the primary topological evolution during representation learning is an increase in point-cloud connectivity, while higher-order topological structures remain relatively limited throughout optimization.
\end{itemize}

\begin{figure*}[!t]
    \centering
    \includegraphics[width=1.0\linewidth]{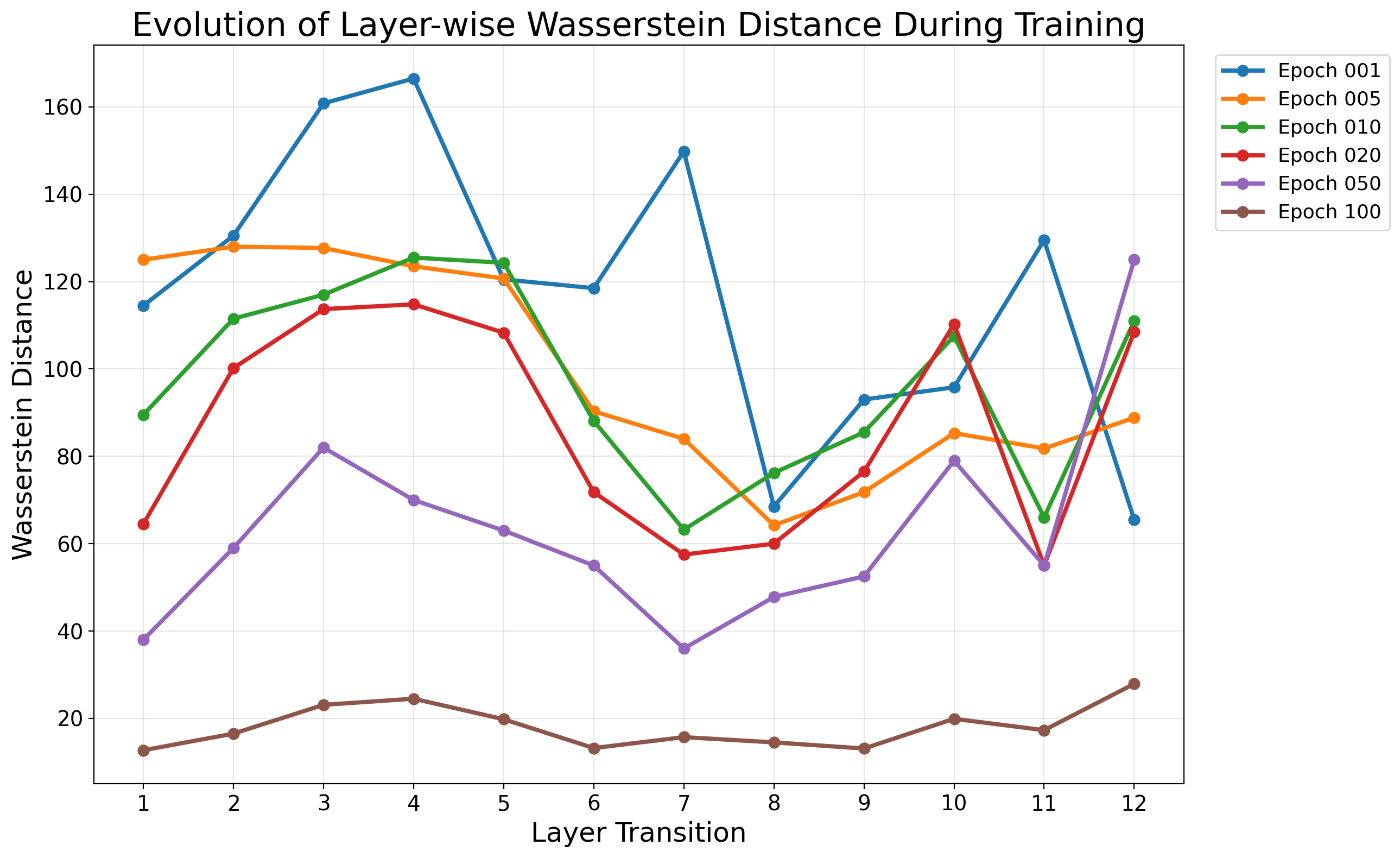}
\caption{Evolution of the Wasserstein Distance between persistence diagrams of consecutive Transformer layers at representative training epochs. Unlike the Bottleneck Distance, the Wasserstein Distance measures the cumulative difference across all persistent topological features. The overall decrease throughout training suggests progressively more similar global persistent topological signatures between adjacent Transformer layers, while localized peaks indicate layer transitions that continue to exhibit comparatively larger cumulative topological reorganization.}
    \label{fig:wasserstein_distance}
\end{figure*}

\subsection{Wasserstein Distance Evolution}

While the Bottleneck Distance measures the largest topological discrepancy between consecutive Transformer layers, the Wasserstein Distance quantifies the cumulative difference between their persistence diagrams. Consequently, it captures the overall evolution of persistent topological features rather than only the most significant structural change. Figure~\ref{fig:wasserstein_distance} illustrates the layer-wise Wasserstein Distance throughout training. Similar to the Bottleneck Distance, the overall magnitude decreases progressively as optimization proceeds. The cumulative topological difference between consecutive layers is largest during the early stages of training and steadily reduces towards convergence, suggesting that neighbouring layers gradually develop increasingly similar persistent topological signatures.

Unlike the Bottleneck Distance, the Wasserstein Distance exhibits a comparatively smoother evolution across layer transitions, indicating that the overall topological structure changes more gradually than the largest individual topological event alone would suggest. Nevertheless, several transitions, particularly around the early encoder blocks and the final Transformer layers, consistently exhibit comparatively larger Wasserstein distances throughout training, implying that these stages continue to contribute more substantially to the cumulative topological evolution of the representation point cloud. Overall, the progressive reduction in Wasserstein Distance suggests that the global persistent topology of neighbouring Transformer layers becomes increasingly similar throughout optimization. Consistent with the methodological considerations discussed in Section~III, these observations should be interpreted as comparative changes in the persistent topology of the constructed simplicial complexes rather than exact measurements of the underlying representation manifold.

\section{Hypothesis: Progressive Topological Stabilization}

The topological observatories presented in TGO-IV collectively suggest a consistent developmental pattern throughout Transformer optimization. Individually, each observable admits multiple interpretations; however, when considered together, they support a coherent explanation of how the representation topology evolves during learning. The progressive leftward shift of the $\beta_0$ curves indicates that connected components merge at increasingly smaller filtration radii as training proceeds. In isolation, this observation admits two competing interpretations: (i) a trivial collapse of the representation point cloud into an increasingly featureless structure, or (ii) the emergence of a more compact yet organized representation. Consequently, the evolution of $\beta_0$ alone is insufficient to distinguish between these possibilities. The persistent presence of non-trivial $H_1$ features throughout training argues against complete topological degeneration. Although the number and persistence of loops remain comparatively limited, they consistently survive throughout optimization, indicating that the representation does not collapse into a topologically trivial point cloud. Instead, the learned representation preserves non-trivial geometric organization even as its global connectivity increases. A complementary perspective is provided by the higher-order homology. The transient emergence of $H_2$ features during intermediate stages of training, followed by their gradual attenuation towards convergence, suggests that higher-order geometric structures are actively formed during optimization before being progressively simplified. Rather than indicating continuous topological collapse, this behaviour is more consistent with a process of geometric restructuring followed by refinement.

Finally, the Wasserstein Distance exhibits a substantial reduction throughout training while largely preserving its characteristic depth-wise profile. Since the Wasserstein Distance measures the cumulative discrepancy between persistence diagrams, this observation suggests that successive Transformer layers induce progressively smaller topological modifications as optimization converges, rather than fundamentally altering the developmental pattern of representation evolution. Taken together, these observations motivate the following hypothesis.

\begin{quote}
\textbf{Progressive Topological Stabilization Hypothesis.}
Transformer representations evolve through a process of progressive topological stabilization, in which the representation point cloud becomes increasingly compact while preserving non-trivial geometric organization. During optimization, higher-order topological structures are transiently established and subsequently refined, whereas successive Transformer layers perform progressively smaller cumulative topological transformations as convergence is approached.
\end{quote}

It is important to emphasize that this hypothesis is intended as an empirical interpretation of the persistent topological observables obtained in TGO-IV. Owing to the sparse point-cloud approximation used to construct the Vietoris--Rips complexes, the proposed hypothesis should be interpreted as describing the evolution of the persistent topology of the constructed simplicial complexes rather than the intrinsic topology of the underlying representation manifold. Future observatories employing denser point-cloud approximations and larger-scale topological analyses may further validate or refine this hypothesis.

\section{Conclusion}

This work introduced \textbf{TGO-IV: Developmental Topology Observatory}, a topological framework for investigating the developmental evolution of Transformer representations through the lens of Persistent Homology. By constructing Vietoris--Rips simplicial complexes from layer-wise representation point clouds, TGO-IV characterized representation development using complementary topological observatories including Persistence Diagrams, Barcode Diagrams, Betti Curves, Persistence Landscapes, Bottleneck Distance, and Wasserstein Distance. The empirical analysis revealed several consistent developmental trends throughout optimization. Connected components merged at progressively smaller filtration radii, persistent one-dimensional topological structures were preserved throughout training, higher-order topological features emerged transiently before gradually simplifying, and the cumulative topological discrepancy between consecutive Transformer layers steadily decreased as optimization converged. Collectively, these observations suggest that representation learning is accompanied by increasingly stable persistent topological signatures rather than arbitrary topological evolution.

Motivated by these observations, TGO-IV proposed the \emph{Progressive Topological Stabilization Hypothesis}, which posits that Transformer representations evolve through geometric refinement toward increasingly compact yet non-trivial topological organizations. Rather than supporting an interpretation of trivial representational collapse, the observed persistent homological signatures are more consistent with a process in which higher-order geometric structures are established, refined, and progressively stabilized throughout learning. As the fourth observatory in the Transformer Geometry Observatory (TGO) framework, TGO-IV extends the study of Transformer representations beyond spectral, geometric, and semantic analyses by providing a complementary topological perspective on representation development. Together with the preceding observatories, these results contribute toward a more comprehensive understanding of how Transformer representations evolve throughout the learning process.

\section{Future Works}

The preceding observatories within the \emph{Transformer Geometry Observatory} have progressively expanded the study of Transformer representations from spectral geometry (TGO-I), representation geometry (TGO-II), semantic geometry (TGO-III), and persistent topology (TGO-IV). Collectively, these observatories demonstrate that Transformer representations undergo highly structured developmental evolution across multiple complementary perspectives. While these studies establish consistent empirical patterns, they do not yet explain the computational mechanisms responsible for their emergence.

Consequently, the remaining observatories shift their focus from describing representation evolution to understanding the computational and optimization processes that give rise to it. Rather than introducing additional geometric observables, the subsequent studies aim to investigate how Transformer computations, gradient dynamics, and optimization landscapes collectively produce the developmental behaviours consistently observed throughout the previous TGOs.

\subsection{TGO-V: Computational Geometry Observatory}

While TGO-IV reveals that Transformer representations undergo progressive topological stabilization throughout training, it does not explain how individual Transformer computations generate these topological transformations. TGO-V will therefore investigate token trajectories, representation drift, attention routing, information flow, and local computational geometry throughout the forward pass. These experiments aim to identify the computational mechanisms responsible for the progressive geometric and topological refinement observed throughout the previous observatories.

\subsection{TGO-VI: Optimization Dynamics Observatory}

Having established the computational evolution of Transformer representations, the final observatory will investigate the optimization principles governing this developmental process. TGO-VI will analyze gradient dynamics, Hessian spectra, curvature evolution, optimization trajectories, singular learning theory, and loss landscape geometry. These experiments aim to determine how optimization progressively shapes the spectral, geometric, semantic, and topological organization observed throughout Transformer learning.

\subsection{Long-Term Objective}

The long-term objective of the Transformer Geometry Observatory is to establish a unified mechanistic theory describing how Transformers learn. By integrating spectral geometry, representation geometry, semantic geometry, persistent topology, computational dynamics, and optimization dynamics, the complete framework seeks to bridge empirical observations with the underlying principles governing representation learning. Ultimately, this unified understanding may provide theoretical insights for the design of more interpretable, efficient, and biologically-inspired learning systems.

\bibliographystyle{unsrt}
\bibliography{references}
\end{document}